\documentclass[sigconf]{acmart}

\usepackage{subcaption}
\usepackage{mathtools}

\DeclarePairedDelimiter\floor{\lfloor}{\rfloor}

\usepackage{paralist}

\usepackage[]{natbib}
\usepackage{xcolor, colortbl}

\definecolor{red}{rgb}{0.894, 0.102, 0.11}
\definecolor{blue}{rgb}{0.216, 0.494, 0.722}
\definecolor{green}{rgb}{0.302, 0.686, 0.29}
\definecolor{purple}{rgb}{ 0.596, 0.306, 0.639}
\definecolor{orange}{rgb}{ 1.00, 0.498, 0}
\definecolor{grey}{rgb}{0.60, 0.060, 0.60}
\definecolor{brown}{rgb}{0.651, 0.337, 0.157}
\definecolor{pink}{rgb}{0.969, 0.506, 0.749}

\newcommand{\todo}[1]{{\leavevmode\color{red}#1}}
\renewcommand{\todo}[1]{}
\newcommand{\com}[1]{{\leavevmode\color{blue}#1}}
\renewcommand{\com}[1]{}
\newcommand{\added}[1]{{\leavevmode\color{green}#1}}
\renewcommand{\added}[1]{#1}
\newcommand{\comMo}[1]{{\leavevmode\color{orange}#1}}
\renewcommand{\comMo}[1]{}
\newcommand{\todoMo}[1]{{\leavevmode\color{purple}#1}}
\renewcommand{\todoMo}[1]{}
\newcommand{\raus}[1]{{\leavevmode\color{gray}#1}}
\renewcommand{\raus}[1]{}

\newcommand{\cb}[2]{{\fcolorbox{#1}{#1!0}{#2}}}

\AtBeginDocument{%
  \providecommand\BibTeX{{%
    \normalfont B\kern-0.5em{\scshape i\kern-0.25em b}\kern-0.8em\TeX}}}

\setcopyright{acmcopyright}
\copyrightyear{2019}
\acmYear{2019}

\acmConference[epiDAMINK '19]{epiDAMINK 2019: Epidemiology meets Data Mining and Knowledge discovery, Workshop held in conjuction with ACM SIGKDD 2019}{August 05, 2019}{Anchorage, Alaska - USA}

\begin{document}

\title[Improving Outbreak Detection with Stacking of Surveillance Methods]{Improving Outbreak Detection with \\ Stacking of Statistical Surveillance Methods}

\author{Moritz Kulessa}
\affiliation{%
  \institution{Knowledge Engineering Group\\ Technische Universit{\"a}t Darmstadt}
  \country{Germany}
}
\email{mkulessa@ke.tu-darmstadt.de}
\author{Eneldo Loza Menc\'{i}a}
\affiliation{%
  \institution{Knowledge Engineering Group\\ Technische Universit{\"a}t Darmstadt}
  \country{Germany}
}
\email{eneldo@ke.tu-darmstadt.de}
\author{Johannes F{\"u}rnkranz}
\affiliation{%
  \institution{Knowledge Engineering Group\\ Technische Universit{\"a}t Darmstadt}
  \country{Germany}
}
\email{juffi@ke.tu-darmstadt.de}

\begin{abstract}
  Epidemiologists use a variety of statistical algorithms for the early detection of outbreaks. The practical usefulness of such methods highly depends on the trade-off between the detection rate of outbreaks and the chances of raising a false alarm. Recent research has shown that the use of machine learning for the fusion of multiple statistical algorithms improves outbreak detection. Instead of relying only on the binary output (\emph{alarm} or \emph{no alarm}) of the statistical algorithms, we propose to make use of their $p$-values for training a fusion classifier. In addition, we also show that adding additional features and adapting the labeling of an epidemic period may further improve performance. 
  For comparison and evaluation, a new measure is introduced which captures the performance of an outbreak detection method  with respect to a low rate of false alarms more precisely than previous works.
  Our results on synthetic data show that it is challenging to improve the performance with a trainable fusion method based on machine learning. In particular, the use of a fusion classifier that is only based on binary outputs of the statistical surveillance methods can make the overall performance worse than directly using the underlying algorithms. However, the use of $p$-values and additional information for the learning
  is promising, enabling to identify more valuable patterns to detect outbreaks.

  
\end{abstract}

\maketitle

\section{Introduction}
\label{section:background}


The early detection of infectious disease outbreaks is of great significance for public health. 
The spread of such outbreaks could be diminished tremendously 
by applying control measures as early as possible, which indeed can save lives and reduce suffering~\cite{FarringtonImproved}.
For that purpose, statistical algorithms have been developed to automate and improve outbreak detection. Such methods raise alarms in the case that an unusually high number of infections is detected which results in a further investigation by an epidemiologist~\cite{odm_compare3}.
Ideally, such algorithms are completely automated while still being able to be applied on a wide spectrum of different infections and syndromes~\cite{odm_compare_noufaily}. However, if not chosen wisely or configured properly, they may also raise many false alarms which can overwhelm the epidemiologist. In particular for large surveillance systems, where many time series for different diseases and different locations are monitored simultaneously, the false alarm rate is a major concern and therefore highly determines the practical usefulness of an outbreak detection method~\cite{statistical_challenges}. However, regulating the false alarm rate usually has an impact on the ability to detect outbreaks. To find a good trade-off between those measures is one of the major challenges in outbreak detection~\cite{FarringtonImproved, odm_compare2}.

Traditional outbreak detection methods rely on historic data to fit a parametric distribution which is then used to check the statistical significance of the current observation. Choosing the significance level for the statistical method beforehand makes the evaluation difficult. In line with \citet{measures1}, we propose a method which uses the $p$-values of the statistical methods in order to evaluate their performance. 
\raus{By doing so we are able to precisely measure the trade-off between the false alarm rate and the detection rate.} 
In particular, we propose a variant of Receiver Operating Characteristic (ROC) curves, which shows the false alarm rate on the $x$-axis and the detection rate---in contrast to the true positive rate---on the $y$-axis. By using the area under the \emph{partial} ROC curve~\citep{partial_ROC}, we are able to obtain a measure for the performance of an algorithm satisfying a particular constraint on the false alarm rate (e.g. less than 1\% false alarms). This criterion serves as the main measure for our evaluations and enables to analyze the trade-off between the false alarm rate and the detection rate of outbreak detection methods precisely.

Prior work on outbreak detection mainly focuses on forecasting the number of infections for a disease (e.g. \cite{forecasting1, forecasting2}). However, only little research has been devoted to use supervised machine learning~(ML) techniques for improving algorithms, which can raise alarms.
\citet{surveillance_ml} used \emph{Baysian networks} to identify the determinants for detection performance to find appropriate algorithm configurations for outbreak detection methods. 
Furthermore, classification algorithms and voting schemes have been used for the fusion of outbreak detection methods on univariate time series~\cite{fusion1, fusion2} as well as on  multi-stream time series~\cite{fusion_multi_1, fusion_multi_2, fusion_evidence}.
However, the examined approaches only rely on the binary output (\emph{alarm} or \emph{no alarm}) of the underlying statistical methods for the fusion which limits the information about a particular observation. Prior research in the area of ML has shown that more precise information of the underlying models improves the overall performance of the fusion~\cite{stacking_issue}. Therefore, we propose an approach for the fusion of outbreak detection methods which uses the $p$-values of the underlying statistical methods. Moreover, one can also incorporate different information for the outbreak detection (e.g., weather data, holidays, statistics about the data, \ldots) by just augmenting the data with additional attributes. 
As a first step, we put our focus on improving the performance of outbreak detection methods using an univariate time series as the only source of information. 
Furthermore, the way outbreaks are labeled in the data also has a major influence on the learnability of outbreak detectors. Thus, we propose adaptions for the labeling of outbreaks in order to maximize the detection rate of ML algorithms.


\section{Statistical Algorithms for Syndromic Surveillance}
\label{sec:algorithms}

The key idea of our approach is to learn to combine predictions of commonly used statistical outbreak detection methods with a trainable ML algorithm. Thus, we first need to generate a series of aligned prediction vectors, each consisting of one entry for each method. This sequence can then be used for training the ML model.

Let us denote with $C=(c_0, c_1, \ldots, c_n) \in \mathbb{N}^n$ the time series of infection counts for a particular disease. 
Many methods rely on a sliding window approach which uses the previous $m$ counts as reference values for fitting a particular parametric distribution. Therefore, the mean $\mu(t)$ and the variance $\sigma^2(t)$ can be computed over these $m$ reference values as follows:
\begin{align*}
\mu(t) = \frac{1}{m} \sum_{i=1}^{m} c_{t-i} && \sigma^2(t) = \frac{1}{m} \sum_{i=1}^{m} (c_{t-i} - \mu )^2
\end{align*}
On the fitted distributions, a statistical significance test is performed in order to identify suspicious spikes. 
\added{For the purpose of outbreak detection, we rely on one tailed-tests for the statistical algorithms in order to only capture the observation of unusual high number of infections. For a particular observed count $c_t$ and a fitted distribution $p(x)$,
the $p$-value is computed as the probability $\int_{c_t}^{\infty} p(x) dx$ of observing $c_t$ or higher counts. Hence, small $p$-values represent uncommonly high counts of $c_t$.}
The sensitivity of raising an alarm is regulated by the significance level $\alpha$ and if the $p$-value is inferior to the threshold $\alpha$ an alarm is raised. 

We have chosen to base our work on the following methods 
which are all implemented in the R package \textit{surveillance}~\cite{surveillanceR}:


\begin{description}

\item[EARS C1] and \textbf{EARS C2} are variants of the \emph{Early Aberration Reporting System}  \citep{hutwagner03EARS,fricker08EARS} which rely on the assumption of a Gaussian distribution. 
The difference between C2 and C1 lies in the added gap of two time points between the reference values and the current observed count $c_t$, so that the distribution of $c_t$ are assumed as in the following:
\begin{equation*}
c_t \stackrel{\text{C1}}{\sim} N(\mu(t), \sigma^2(t))
\qquad
c_t \stackrel{\text{C2}}{\sim} N(\mu(t-2),\  \sigma^2(t-2))
\end{equation*}\com{only saved some lines}

\item[EARS C3] combines the result of the C2 method over a period of three previous observations. For convenience of notation, the incidence counts $c_t$ for the C3 method are transformed according to the statistics so that it fits to the normal distribution.
\begin{align*}
\left[\frac{c_t -\mu(t-2)}{\sqrt{\sigma^2(t-2)}}  - \sum_{i=1}^2 \max(0, \frac{c_{t-i} -\mu(t-2-i)}{\sqrt{\sigma^2(t-2-i)}}  - 1)\right] \stackrel{\text{C3}}{\sim} N(0,1)
\end{align*}
Despite the inaccurate assumption of the Gaussian distribution for low counts, the EARS variants are often included in comparative studies due to its simplicity and still serves as competitive baseline \citep{odm_compare2,fricker08EARS,odm_compare}.

\item[Bayes method.] In contrast to the family of C-algorithms, the Bayes algorithm relies on the assumption of a negative binomial distribution:
\begin{equation*}
c_t \stackrel{\text{Bayes}}{\sim} NB(m \cdot \mu(t) + \frac{1}{2},\  \frac{m}{m+1})
\end{equation*}

\item[RKI method.] Since the Gaussian distribution is not suitable for count data with a low mean, the RKI algorithm, as implemented by \citet{surveillanceR}, assumes a Poisson distribution:
\begin{align*}
c_t \stackrel{\text{RKI}}{\sim}
\begin{cases}
    Poisson(\floor*{\mu(t)}+1),& \text{if } \mu(t) \leq 20 \\
    N(\mu(t),\  \sigma^2(t)),& \text{otherwise}
\end{cases}
\end{align*}
\end{description}

They all have in common that they require comparably little historic data on their own, which allows us to train the ML method on longer sequences. Moreover, such methods are universally applicable and serve as drop-in approaches for surveillance systems since they only rely on the detection of a local increase in incidents without the need to capture effects like seasonality and trend.

\section{Fusion Methods}
\label{sec:fusion}

The combination of information from several sources in order to obtain a unified picture is known as \emph{fusion}~\cite{fusion_general}. 
\emph{Classifier fusion} is a special case which combines the outputs of multiple classifiers in order to improve classification performance. 
In our context, the statistical algorithms for syndromic surveillance can be seen as classifiers, each  classifying the current observation into the classes \textsl{alarm} or \textsl{no alarm}. 
A straight-forward way for combining the predictions of multiple outbreak detection methods is to simply vote and follow the majority prediction. A more sophisticated approach consists of training a classifier that uses the predictions of the detection methods as input, and is trained on the desired output, a technique that is known in ML as \emph{stacking}~\cite{first_stacking}. 


\begin{figure*}[t]
    
    \begin{minipage}{0.4805\linewidth}
        \centering
		\includegraphics[width=0.99\linewidth]{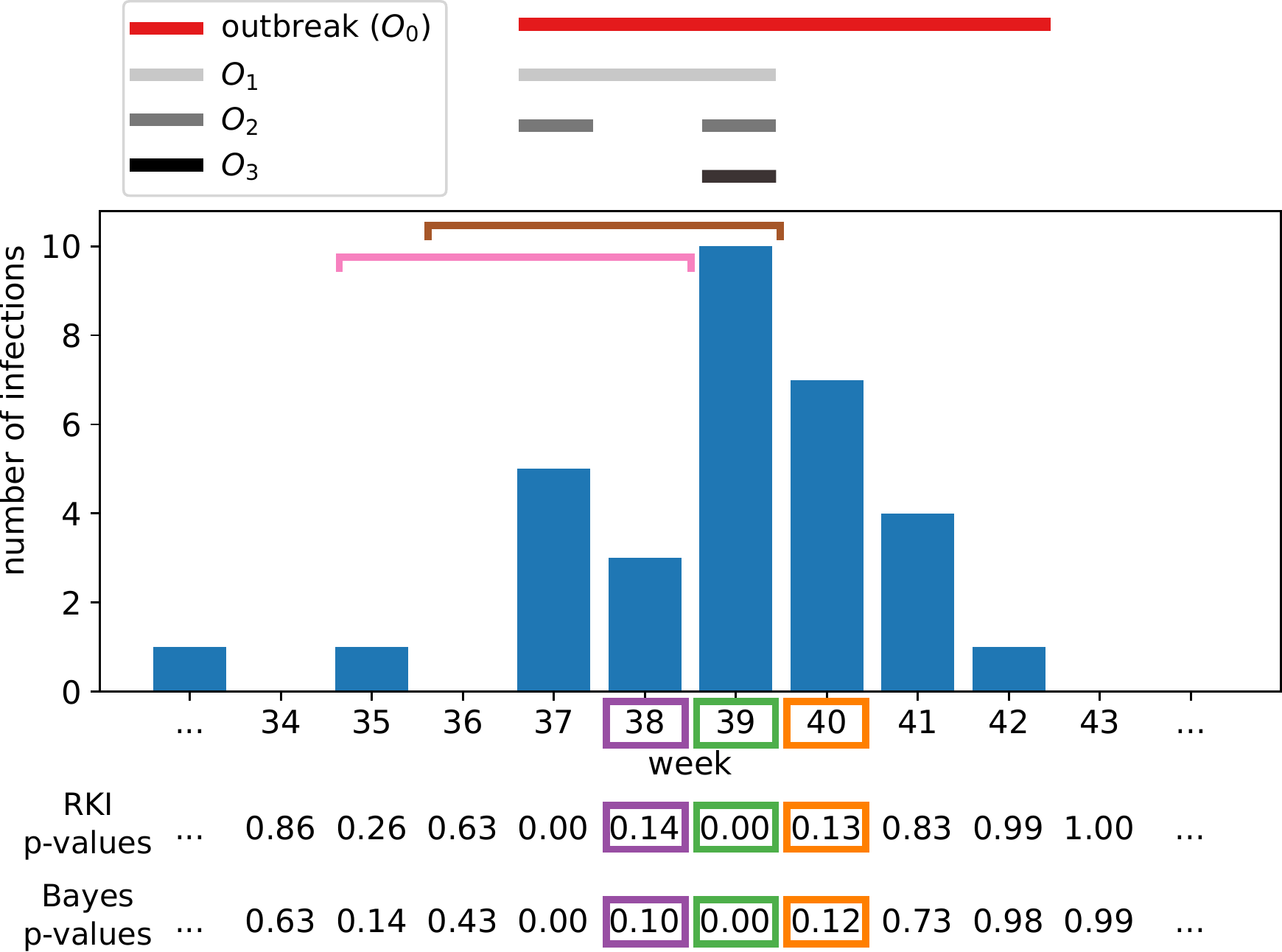}
	\end{minipage}\hfill
	\begin{minipage}{0.48\linewidth}
		\centering
\resizebox{\textwidth}{!}{
		
		\begin{tabular}{c|ccccc|c}
\toprule
index & \multicolumn{3}{c|}{augmented features} & \multicolumn{2}{c|}{$p$-values}\\
week & \multicolumn{1}{c|}{} & \multicolumn{2}{c|}{prev. $p$-values} &  & \multicolumn{1}{c|}{} & target\\
$t$ & \multicolumn{1}{c|}{$\text{mean}_{t}$} & $\text{RKI}_{t-1}$  & \multicolumn{1}{c|}{$\text{Bayes}_{t-1}$} & $\text{RKI}_{t}$ & $\text{Bayes}_{t}$ & $\text{outbreak}_{t}$\\
\midrule
\dots &\dots & \dots & \ldots & \dots & \dots & \ldots\\
34 & 1.00 & 0.59 & 0.63& 0.86  & 0.63 & no\\
35 & 0.50 & 0.86 & 0.63& 0.26  & 0.14 & no\\
36 & 0.50 & 0.26 & 0.14& 0.63  & 0.43 & no\\
37 & 0.50 & 0.63 & 0.43& 0.00  & 0.00 & \cb{red}{yes}\\
\cb{purple}{38} & 1.50 & 0.00 & 0.00& \cb{purple}{0.14}  & \cb{purple}{0.10} & \cb{red}{yes}\\
\cb{green}{39} & \cb{pink}{2.25} & \cb{purple}{0.14} & \cb{purple}{0.10}& \cb{green}{0.00}  & \cb{green}{0.00} & \cb{red}{yes}\\
\cb{orange}{40} & \cb{brown}{4.50} & \cb{green}{0.00} & \cb{green}{0.00}& \cb{orange}{0.13}  & \cb{orange}{0.12} & \cb{red}{yes}\\
41 & 6.25 & \cb{orange}{0.13} & \cb{orange}{0.12}& 0.83  & 0.73 & \cb{red}{yes}\\
42 & 6.00 & 0.83 & 0.73& 0.99  & 0.98 & \cb{red}{yes}\\
43 & 5.50 & 0.99 & 0.98& 1.00  & 0.99 & no\\
\ldots & \dots & \dots & \ldots & \dots & \dots & \ldots\\
\bottomrule
\end{tabular}
}
	\end{minipage}
\captionof{figure}{Example for the creation of training data for the learning algorithm including  the statistical algorithms Bayes and RKI with a window size of one ($w=1$) and the mean over the previous four counts ($m=4$) as features. On the left hand side, the time series for a particular disease is visualized at the center representing the number of cases of infections over time.
The computed $p$-values of the statistical algorithms (underneath) and the label indicating an outbreak for each observation (above) are placed at the respective time index $t$. Using this information the data instances can be created as shown on the right: Each particular time point is represented by one training instance,
labeled according to the original targets $O_0$.
}
\label{figure:example_dataset_creation}
\end{figure*}

Recent work in the area of outbreak detection and fusion has focused on fusing the information obtained by simultaneously monitoring multiple time series for a particular disease. \citet{fusion_multi_1} have shown that the performance of statistical algorithms 
can already be improved by combining them with simple voting schemes. \citet{fusion_multi_2} could further improve the performance using Bayesian networks and including further information about the patients (e.g., age) as additional attributes. Moreover, \citet{fusion_evidence} have used a hierarchy of Bayesian networks in order to incorporate additional information about health surveillance data and environmental sensors. However, all of these fusion methods aim to capture the degree of dependence between the monitored time series relying on spatial correlations.

Only little research has been devoted to improving the performance of statistical algorithms on univariate time series. In particular, \citet{fusion1} have used the ML technique \emph{hierarchical mixture of experts} \cite{HME} to combine the output of the methods from EARS.
However, the authors note that all algorithms rely on the assumption of a Gaussian distribution, which limits their diversity.
In contrast, \citet{fusion2} have used a variety of classification algorithms (\emph{logistic regression}, \emph{CART} and \emph{Baysian Networks}) for the fusion of outbreak detection methods. As underlying statistical algorithms they have used the Cumulative Sum (CUSUM), two Exponential Weighted Moving Average algorithms, the EARS methods (C1,C2,C3) and the Farrington algorithm~\cite{FarringtonImproved}.
In general, the results of \citet{fusion1} and \citet{fusion2} indicate that ML improves the ability to detect outbreaks while simple voting schemes (e.g. weighted voting and majority vote) did not perform well. Moreover, the algorithms have not been evaluated with respect to data which include seasonality and trend.

\section{Fusion with Augmented Stacking}
\label{sec:proposed}

In this work, we show that the availability of additional information can further improve the performance of the fusion classifier. 
Therefore, we first propose to use $p$-values of the statistical methods for the fusion in order to include information about the certainty of an alarm, 
and then show how to add additional external information to the learning process of the ML algorithm. Finally, we investigate different variants for labeling outbreaks.

\subsection{Fusion with $p$-values}
\label{sec:p-values}



Given base estimators $g_1(x), \ldots, g_K(x)$, a \emph{fusion combiner} is a function $h(g_1(x),\ldots, g_K(x))$ that combines the predictions of the base functions. In the simple case of binary voting, i.e., $g_i(x) \in \{0,1\}$,  the combiner $h(x)=\frac1K \sum_i g_i(x)$  with a threshold of $0.5$ would model the majority rule.
In \emph{stacking} the function $h: X^K \xrightarrow{} O$ is learned by 
training a machine learning classifier on a set of previous observations $(g_1(x_1), \ldots, g_K(x_1))$, \ldots, $(g_1(x_n), \ldots, g_K(x_n))$ --derived from applying $g_i$ on $x_t$--
with associated targets $o_1, \ldots, o_n \in O$. We refer to this as the training set in contrast to the evaluation set, which contains new, unseen observations. 
In outbreak detection, the instances $x_t$ correspond to the points in the time series $C$ of infection counts $c_t$ and 
$o_t \in \{0,1\}$ denotes the labelling of a time point as belonging to an outbreak (1) or not (0).

Previous approaches \citep{fusion1,fusion2} used the binary alarms (\{0,1\}) of base outbreak detectors. 
In this work instead, we propose to base our stacking model on the $p$-values, i.e., $g_i(x) \in [0,1]$, provided by the underlying  statistical approaches (cf. Sec.~\ref{sec:algorithms}). 
In fact, the $p$-values can directly be seen as the certainty of currently observing an outbreak, enabling the learning algorithm to make use of the base estimations in a much more fine grained way.
This information is otherwise lost when using binary alarms, which are indeed obtained by just applying a fixed threshold on the computed $p$-values.
In addition to the circumvented difficulty of tuning such threshold, previous studies on stacking have shown empirically that using the raw predictions can improve over the discretized option \citep{stacking_issue}.\todo{\@Juffi, is that the paper you mentioned??}

Figure~\ref{figure:example_dataset_creation} visualizes an example on how the data for the learning algorithm is created by using the $p$-values of the statistical algorithms Bayes and RKI. The columns RKI$_t$ and Bayes$_t$ represent the computed $p$-values for the current observation while the 
other columns (mean$_t$, RKI$_{t-1}$ and Bayes$_{t-1}$) represent additional information explained in the following section.

\subsection{Additional Features}
\label{sec:additional_features}
The use of a trainable fusion method allows us to include additional information which can help to decide whether a given alarm should be raised or not.
As additional features, we propose to include the \emph{mean} of the counts over the last $m$ time points (the same number of time points as used by the statistical methods), which can give us evidence about the reliability of a particular outcome. For example, the assumption of a Gaussian distribution for a low mean of count data ($\leq20$) is known to be imprecise. Therefore, a learning algorithm might induce in this scenario that the $p$-values of the statistical methods C1, C2 and C3 may not be trustworthy. Moreover, under the assumption that a time series is stationary an unusual high mean can also be a good indicator to detect an outbreak, especially in the case that an outbreak arises slowly over time. 
The column $\text{mean}_{t}$ in Figure~\ref{figure:example_dataset_creation} illustrates how the mean over the last four observed counts ($m=4$) is added as an additional feature.

Finally, we also include the output of the statistical methods for previous time points in a window of a user-defined size $w$ as additional features. 
\raus{Cases where no (previous) output for an outbreak detection method exists are denoted with missing values, which can be adequately used 
by a wide variety of ML algorithms.
Moreover, the use of missing values also has the advantage that we are able to learn and predict outbreaks for scenarios in which only a subset of the underlying statistical algorithms can provide an output. For example, if the underlying methods rely on different amount of historic data, not all methods are able to provide an output for the first observations.} 
For the example in Figure~\ref{figure:example_dataset_creation}, we have used a window size of one ($w=1$) which includes the previous output of both statistical algorithms.



\subsection{Modelling the Output Labels for Learning}
\label{sec:adaption_labeling}
A major challenge for ML algorithms is that the duration of an outbreak period is not clearly defined~\cite{statistical_challenges}. A simple strategy---which we refer to as $O_0$---is to label all time points positive as long as cases for the particular epidemic are reported (e.g. time points prior to the peak of an outbreak and a few time points after the peak). 
In this case, the goal of the learning algorithm is to predict most time points in an ongoing epidemic as positive, regardless of their time stamp.
Indeed, our early results indicate that the predictor learns to recognize the fading-out of an outbreak (e.g. weeks 40 to 42 in Figure~\ref{figure:example_dataset_creation}). 
This is due to the fact that the peak of the outbreak is included in the reference values which results in a considerably high mean $\mu(t)$ for the significance test. Because of this, unusually high $p$-values are generated for the counts after the peak, which provide sufficient evidence for the stacking algorithm to raise an alarm.
However, this also increases the number of false alarms as the ML approach learns to raise alarms when the count is decreasing outside an epidemic period. 

To avoid this, we propose three adaptations of $O_0$: $O_1$ labels all time points until the peak (the point with maximum number of counts during the period) as  positive. $O_2$ instead skips the time points whose count is decreasing compared to the immediate previous count (i.e., it labels all increasing counts until reaching the peak). Finally, $O_3$ labels only the peak of the outbreak as positive.
Figure~\ref{figure:example_dataset_creation} visualizes an example outbreak with the corresponding different options to label the epidemic period on the top-left.


\section{Evaluation Measures}
\label{sec:measures}



Instead of manually adjusting the $\alpha$ parameter of the statistical methods and examining the results individually, which is mostly done in previous works, we propose to evaluate the $p$-value as it is done by \citet{measures1}. In particular, the $p$-value can be interpreted as a score, which sorts examples according to their degree to which they indicate an alarm. This allows us to
analyze an algorithm with ROC curves~\cite{roc_analysis}. A ROC curve can be used to examine the trade-off between the \emph{true positive rate} (i.e., the probability of raising an alarm in case of an actual outbreak) and the \emph{false alarm rate} (i.e., the probability of falsely raising an alarm when no outbreak is ongoing). In order to only focus on high specificity results (e.g., with a false alarm rate below $1\%$), which is of major importance for many medical applications, we only consider \emph{partial ROC curves}. 
By using the partial area under the ROC curve as proposed in \cite{partial_ROC}, we obtain a simple measure to evaluate the performance of an algorithm, satisfying particular constraint on the false alarm rate. We refer to this measure as \emph{pAUC$_e$} where the parameter $e$ defines the maximum allowed false alarm rate to be considered. It is computed as
\begin{equation*}
\text{\emph{pAUC}}_e = \frac{\int_0^e ROC(f) \ df}{e} 
\end{equation*}
where $ROC(f)$ denotes the true positive rate given a false alarm rate of $f$.
\begin{figure}[tb]
\centering
\includegraphics[width=0.90\linewidth]{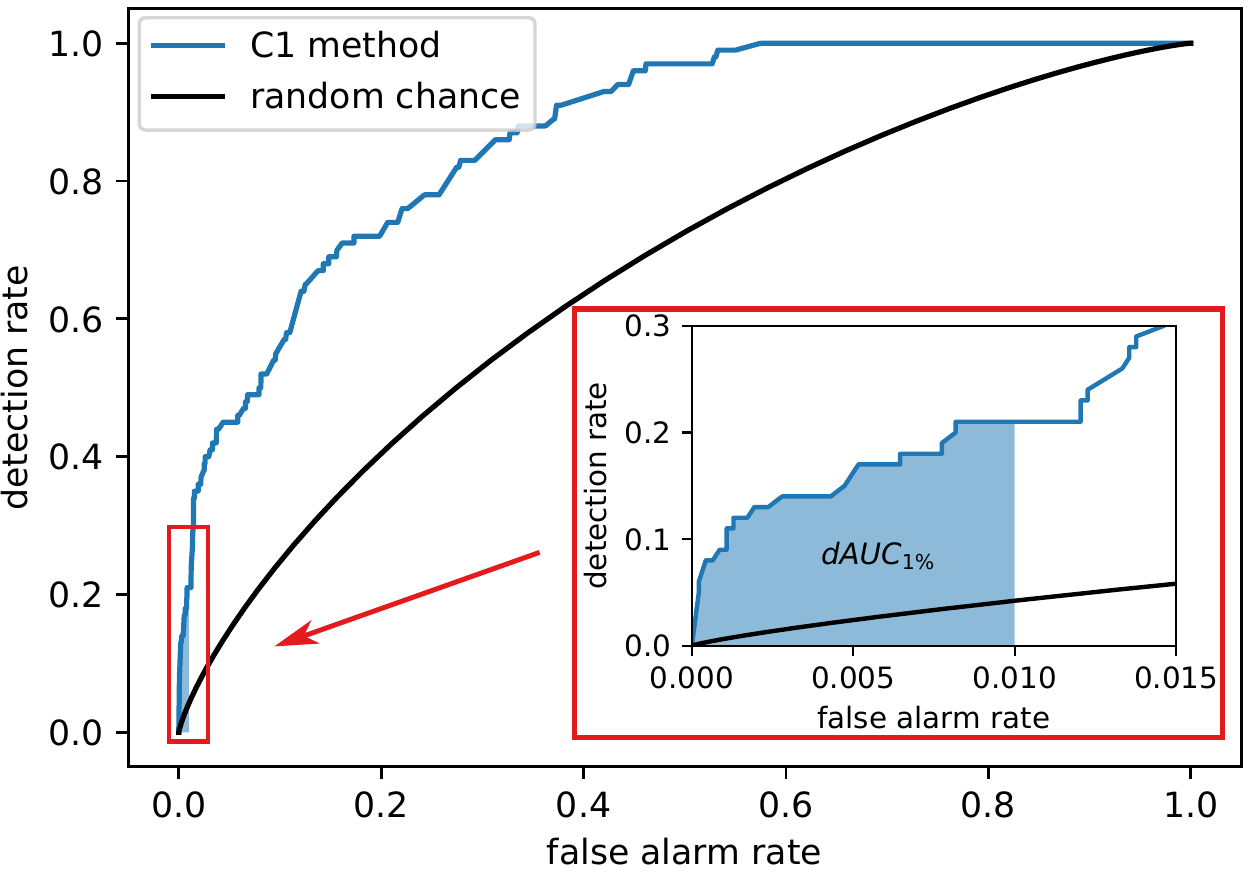}
\caption{ROC curve using the detection rate on the $y$-axis. 
The better-than-chance performance is lifted above the diagonal since the detection rate is an interval-based metric.
}
\label{figure:dAUC}
\end{figure}
However, alarms raised in cases when the epidemic has already been detected are typically not very decisive and informative anymore. To incorporate this, we consider the \emph{detection rate}, which represents the proportion of recognized outbreaks (i.e., the outbreaks in which at least one alarm is raised during their activity). Following \citet{measures1} and \citet{fusion2}, we therefore use a ROC curve-like representation with the detection rate on the $y$-axis instead of the true positive rate, and use \emph{dAUC$_e$} to refer to the partial area under this curve. Figure~\ref{figure:dAUC} shows an example of the ROC-curve like representation and visualizes the area of \emph{dAUC$_{1\%}$}.\todo{nur persoenlicher geschmack: ich wuerde $\theta$ oder $\tau$ lieber verwenden als $e$.}
%
\added{\citet{measures1} proposed to use weighted ROC curves to also incorporate the influence of the measure timeliness (mean time to detect an outbreak). However, we argue that the weighing with the timeliness introduces a trade-off (importance of timeliness over detection rate)\com{, which is difficult to determine globally,(?)} and a loss in interpretability of the absolute numbers.}\todo{that extra paragraph could be removed}

\section{Evaluation}
\label{section:results}

\added{
The key aspect of our experimental evaluation is to demonstrate that the fusion of $p$-values leads to a further improvement in performance compared to only using the binary output of the statistical algorithms. However, for a deeper understanding of our proposed approaches,\todo{why fair? comprehensive? in-depth analysis, we previously ...?} we first performed experiments to evaluate the influence of including additional features for the stacking in Section~\ref{sec:eval_additional_features}, followed by an analysis of adapting the labeling for the learning in Section~\ref{sec:eval_adaption_labeling}. Finally, using the obtained knowledge about the effect of the proposed techniques, we compare them with the underlying statistical algorithms in Section~\ref{sec:eval_compare}, which represents our main result.
}




\subsection{Experimental Setup}
\label{sec:experimental_setup}
As an implementation baseline for the statistical methods, we have used the R package \textit{surveillance}~\cite{surveillanceR} and adapted the implementation of the methods EARS (C1, C2 and C3), Bayes and RKI in order to also return $p$-values. All methods use the previous $7$ time points as reference values, which is the standard configuration. For the ML part, we rely on the Python library \textit{scikit-learn}~\cite{scikitlearn}. To keep the evaluation simple, we use a
\textit{random forest} classifier.
Basically, it learns an ensemble of randomized decision trees, which has proven to be robust in performance theoretically and practically~\cite{delgado14classifiercomparison, wyner17explainingRF}. 
Each model is composed of 100 decision trees with a minimum number of instances per leaf of $5$ and default settings otherwise.
To allow comparability between the fusion methods, we also evaluated the approach which only combines the binary outputs of the statistical methods as proposed in~\citep{fusion1,fusion2} and which we refer to as the \emph{standard fusion}. Our preliminary experiments have shown that $\alpha=0.5\%$ for the underlying statistical methods performs best for this fusion approach.
For all evaluations, we focused on our proposed evaluation measure $dAUC_{1\%}$ where we fixed the constraint on the false alarm rate to be less than $1$\%.

Our evaluation is based on synthetic data which have been proposed by~\citet{FarringtonImproved}. 
In total $42$ different \emph{test cases} are used which reflect a wide range of application scenarios allowing to analyze the effects of trend~(T), seasonality~(S1) and biannual seasonality~(S2) explicitly. For each parameter configuration $100$ time series are generated, each containing a total of $624$ weeks. Following \citet{FarringtonImproved}, the last 49 weeks of each time series serve as \emph{evaluation data} which include exactly one outbreak whereas the first $575$ weeks contain four outbreaks and represents the so called \emph{baseline data}.\todo{If this is called here baseline data (why?), then the term has to be adapted in the stacking section.}
Each outbreak starts at a randomly drawn week and the number of cases per outbreak is generated with a Poisson distribution with the mean equal to a constant $k$ times the standard deviation of the counts observed at the starting week. The outbreak cases are then distributed over time using a log-normal distribution with mean $0$ and standard deviation $0.5$. 
We evaluated each stacking configuration separately for each test case using the baseline data of the 100 time series for training (in total $57.500$ weeks including $400$ outbreaks) and the remaining $4.900$ weeks for testing ($100$ outbreaks), respectively. The statistical methods were applied separately for each time series in order to obtain the $p$-values as inputs for the learner as well as the predictions on the evaluation set.



Instead of reporting the average over $dAUC_{1\%}$ scores, which could have different scales for different test cases, we determined a ranking over the compared methods for each test case. Afterwards, each method's rank is averaged across the evaluated test cases to obtain an overall rank. In order to evaluate the effects of trend and seasonality explicitly, we average the rankings only over the test cases which include these effects. To differentiate between our proposed approaches, we use the notation $M(a,o,w)$ where $M \in \{P,S\}$ specifies whether $p$-value fusion~(P) or the standard fusion~(S) has been used, $a\in \{\neg \mu, \mu\}$ whether the mean is included, $o\in \{O_0, O_1, O_2, O_3\}$ which labeling for the learning, and $w\in \{0,1,\ldots, 12\}$ the window size which has been used for the evaluation. In total, we tested $192$ configurations from which we compare only a small subset 
, respectively,
depending on the analyzed aspect.

\subsection{ Evaluation of Additional Features}
\label{sec:eval_additional_features}

\begin{table*}
\caption{Comparison of including or not including the mean in the data for ML algorithms:  \emph{overall} denotes all $42$ test cases, $\{(\neg)T, (\neg)S1, (\neg)S2\}$ only cases (not) containing trend, annual/biannual seasonality, respectively. Each particular subset, fulfilling constraints on seasonality and trend, include $6$ test cases.}
\label{tab:average}
\begin{tabular}{lrrrrrrr}
\toprule
Approach & Overall & $\{\neg{T}, \neg{S1}, \neg{S2}\}$ &  $\{\neg{T}, S1, \neg{S2}\}$ & $\{\neg{T}, S1, S2\}$ &  $\{ T, \neg{S1}, \neg{S2}\}$ & $\{ T, S1, \neg{S2}\}$ & $\{ T, S1, S2\}$ \\
\midrule
S($\neg \mu$, $O_0$, $0$)  &               3.429 &   3.714 & 3.571 & \textbf{3.000} & \textbf{3.429} & \textbf{3.286} &  3.571 \\
S($ \mu$, $O_0$, $0$)  &               \textbf{3.357} &   \textbf{2.571} & \textbf{3.286} & 3.571 & 3.571 & 3.714 &  \textbf{3.429} \\
\midrule
P($\neg \mu$, $O_0$, $0$)  &               1.905 &   2.571 & 1.857 & 2.000 & 1.714 & 1.714 &  1.571 \\
P($ \mu$, $O_0$, $0$)  &               \textbf{1.310} &   \textbf{1.143} & \textbf{1.286} & \textbf{1.429} & \textbf{1.286} & \textbf{1.286} &  \textbf{1.429} \\
\bottomrule
\end{tabular}
\end{table*}

\begin{table*}
\caption{Comparison of different window sizes for the data (including the mean and using the labeling $O_0$).
}
\label{tab:window}
\begin{tabular}{lrrrrrrr}
\toprule
Approach & Overall & $\{\neg{T}, \neg{S1}, \neg{S2}\}$ &  $\{\neg{T}, S1, \neg{S2}\}$ & $\{\neg{T}, S1, S2\}$ &  $\{ T, \neg{S1}, \neg{S2}\}$ & $\{ T, S1, \neg{S2}\}$ & $\{ T, S1, S2\}$ \\
\midrule
    S($\neg \mu$,$O_0$,0)&               9.738 &   9.000 &  9.571 &  \textbf{8.286} & 11.143 & 10.571 &  \textbf{9.857} \\
  S($\neg \mu$,$O_0$,1) &               \textbf{8.738} &   \textbf{8.857} &  \textbf{7.000} &  9.000 &  \textbf{7.571} &  \textbf{8.857} & 11.143 \\
  S($\neg \mu$,$O_0$,2) &              10.762 &  10.571 & 10.857 & 10.714 & 10.571 & 10.143 & 11.714 \\
  S($\neg \mu$,$O_0$,4) &              11.310 &  11.429 & 11.714 & 11.714 & 10.857 & 12.000 & 10.143 \\
  S($\neg \mu$,$O_0$,6) &              11.619 &  12.714 & 12.286 & 10.286 & 11.571 & 11.857 & 11.000 \\
  S($\neg \mu$,$O_0$,8) &              11.548 &  11.143 & 11.571 & 12.000 & 10.857 & 12.000 & 11.714 \\
   S($\neg \mu$,$O_0$,12) &              11.929 &  12.143 & 12.143 & 13.000 & 11.714 & 11.571 & 11.000 \\
   \midrule
    P($\neg \mu$,$O_0$,0)&               5.000 &   5.714 &  5.000 &  5.714 &  4.429 &  3.714 &  5.429 \\
  P($\neg \mu$,$O_0$,1) &               \textbf{3.405} &   \textbf{3.143} &  \textbf{2.571} &  4.571 &  \textbf{3.286} &  4.286 &  \textbf{2.571} \\
  P($\neg \mu$,$O_0$,2) &               4.381 &   5.000 &  4.714 &  4.000 &  4.571 &  4.571 &  3.429 \\
  P($\neg \mu$,$O_0$,4) &               4.667 &   4.143 &  5.000 &  4.000 &  5.143 &  5.286 &  4.429 \\
  P($\neg \mu$,$O_0$,6) &               4.310 &   5.000 &  4.429 &  3.857 &  3.857 &  3.857 &  4.857 \\
  P($\neg \mu$,$O_0$,8) &               4.000 &   3.000 &  4.000 &  4.714 &  5.000 &  3.857 &  3.429 \\
 P($\neg \mu$,$O_0$,12) &               3.595 &   \textbf{3.143} &  4.143 &  \textbf{3.143} &  4.429 &  \textbf{2.429} &  4.286 \\
\bottomrule
\end{tabular}

\end{table*}

\begin{table*}
\caption{Comparison of the different labeling strategies for the epidemics (not using the average and $w=0$).
}
\label{tab:pre_process}
\begin{tabular}{lrrrrrrr}
\toprule
Approach & Overall & $\{\neg{T}, \neg{S1}, \neg{S2}\}$ &  $\{\neg{T}, S1, \neg{S2}\}$ & $\{\neg{T}, S1, S2\}$ &  $\{ T, \neg{S1}, \neg{S2}\}$ & $\{ T, S1, \neg{S2}\}$ & $\{ T, S1, S2\}$ \\
\midrule
S($\neg \mu$, $O_0$, $0$)  &               6.476 &   6.286 & 5.571 & \textbf{4.857} & 7.143 & 7.571 &  7.429 \\
S($\neg \mu$, $O_1$, $0$)  &               6.738 &   7.286 & 6.714 & 6.286 & 6.429 & 7.000 &  6.714 \\
S($\neg \mu$, $O_2$, $0$)  &               5.738 &   6.286 & 5.714 & 5.286 & \textbf{5.429} & 6.000 &  \textbf{5.714} \\
S($\neg \mu$, $O_3$, $0$)  &               \textbf{5.524} &   \textbf{5.286} & \textbf{5.143} & 5.429 & 6.000 & \textbf{5.429} &  5.857 \\
\midrule
P($\neg \mu$, $O_0$, $0$)  &               3.762 &   3.857 & 3.857 & \textbf{2.714} & 4.857 & 4.000 &  3.286 \\
P($\neg \mu$, $O_1$, $0$)  &               3.262 &   2.857 & 4.143 & 4.857 & 2.429 & 2.429 &  2.857 \\
P($\neg \mu$, $O_2$, $0$)  &               2.690 &   3.143 & 3.000 & 3.286 & 2.714 & 2.143 &  \textbf{1.857} \\
P($\neg \mu$, $O_3$, $0$)  &               \textbf{1.810} &   \textbf{1.000} & \textbf{1.857} & 3.286 & \textbf{1.000} & \textbf{1.429} &  2.286 \\
\bottomrule
\end{tabular}

\end{table*}

\begin{table*}[t!]
\caption{
Comparison of the standard fusion, the $p$-value fusion and each individual statistical algorithm. 
}
\label{tab:best_dr}
\begin{tabular}{lrrrrrrr}
\toprule
Approach & Overall & $\{\neg{T}, \neg{S1}, \neg{S2}\}$ &  $\{\neg{T}, S1, \neg{S2}\}$ & $\{\neg{T}, S1, S2\}$ &  $\{ T, \neg{S1}, \neg{S2}\}$ & $\{ T, S1, \neg{S2}\}$ & $\{ T, S1, S2\}$ \\
\midrule
        C1 &               5.381 &   6.429 & 5.429 & 4.143 & 5.714 & 5.714 &  4.857 \\
        C2 &               4.810 &   4.571 & 4.000 & 4.286 & 5.857 & 5.286 &  4.857 \\
        C3 &               4.690 &   5.429 & 4.571 & 4.286 & 4.857 & 4.429 &  4.571 \\
     Bayes &               2.595 &   4.000 & 3.143 & \textbf{2.571} & \textbf{1.571} & \textbf{1.714} &  2.571 \\
       RKI &               3.619 &   3.571 & 2.857 & 4.571 & 3.714 & 3.286 &  3.714 \\
       \midrule
S($\mu$, $O_3$, $1$)  &               5.238 &   3.000 & 6.000 & 5.714 & 4.714 & 5.857 &  6.143 \\
P($\mu$, $O_3$, $1$)   &               \textbf{1.667} &   \textbf{1.000} & \textbf{2.000} & \textbf{2.429} & \textbf{1.571} & \textbf{1.714} &  \textbf{1.286} \\
\bottomrule
\end{tabular}
\end{table*}

\begin{figure*}[t!]
\centering
\includegraphics[width=1\textwidth]{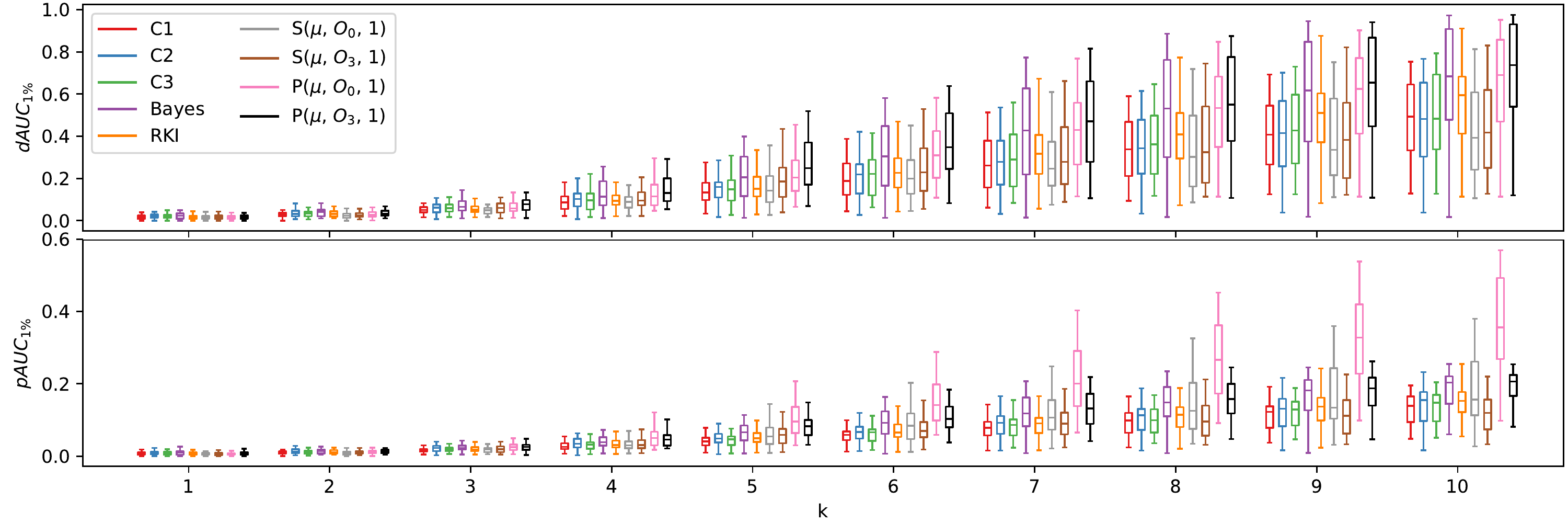}
\caption{Results for the measures \emph{dAUC$_1\%$} and \emph{pAUC$_1\%$}. Each box plot represents the distribution of  measure values for a particular method computed over all $42$ test cases for a fixed outbreak size defined by the parameter $k$ (a bigger value for $k$ indicate more cases per outbreak).}
\label{figure:DRandTPR}
\end{figure*}


The first aspect to review concerns the inclusion of the mean count over the last seven time points. Therefore, we have analyzed the effect of this feature independent of the other parameters using $O_0$ for the labeling of the outbreak and window size $w=0$. The results for the average rank are displayed in Table~\ref{tab:average}.
Comparing the standard to the $p$-value fusion method reveals a beneficial effect especially for the $p$-value approach, for which the variant including the mean achieves an average rank of $1.31$ over $1.91$. In contrast, the average ranks of $3.36$ over $3.43$ for the standard method not only shows that there are issues regarding the usage of the mean for some of the test case configurations, but also the substantial gap between using the binary outputs and the more fine-grained $p$-values. A closer examination reveals that the best improvement for both fusion methods can be achieved on time series without trend and seasonality. By adding effects like trend and seasonality, the mean changes over time, making it difficult for the learning algorithm to use this information. In contrast to the standard fusion, the $p$-value fusion method still enhances by including the mean over the previous time points. \raus{Indeed, our proposed algorithms are not designed to deal with seasonality and trend and we believe incorporating this ability is an interesting avenue for future work.}

The observation that the $p$-value fusion method is superior to the standard fusion can also be seen when comparing different window sizes. The results of this experiment, using $O_0$ for the labeling of the outbreak and not including the mean, are displayed in Table~\ref{tab:window}. In particular, no window configuration of the standard fusion method can outperform any of the $p$-value configurations with respect to the average rank. Overall, a window size of $1$ performed best for both fusion approaches. Being able to compare to the most immediate previous output of the underlying statistical algorithms seems to make it easier to detect anomalies. In contrast, larger window sizes harm the overall performance, which suggests that the additional information is not relevant for detecting sudden changes and rather confuses the learner. Interestingly, on certain combinations of trend and seasonality a larger window size for the $p$-value fusion method seems to be beneficial.
Actually, the increase of the window size also results in taking a further look back in the past allowing to detect effects like trend and seasonality achieving good results on the test cases which only contain biannual seasonality. However, the observed results for larger window sizes are inconsistent across the different test cases, making it difficult to draw valid conclusions.

\subsection{Evaluation of the Labeling Adaptions}
\label{sec:eval_adaption_labeling}

In addition to augmenting the input data, we have evaluated the effect of adapting the labeling of the epidemic period for the training of the stacking algorithm. 
The comparison shown in  Table~\ref{tab:pre_process}  was performed without the augmentation. 

In general,  we can observe that by narrowing the labeling of the outbreak on particular events (i.e., $O_1$, $O_2$ or $O_3$) a better performance can be achieved. 
This effect is clearly visible for the $p$-value fusion method and less obvious for the standard fusion method, for which the adaption $O_1$ seems to be an exception. In particular, learning only the peaks ($O_3$) achieved the best results for both fusion approaches. The benefit of this variant is that the learner can actually focus on the identification of strong and sudden peaks which is indeed the main goal of outbreak detection.
However, in case of biannual seasonality the frequent change of the counts over the season results in many random peaks which apparently makes it difficult for the stacking approach to distinguish between an epidemic peak and a peak caused by random effects.
On the test cases without trend ($\{\neg{T}, S1, S2\}$) outbreaks are better identifiable by also including the fading of the outbreak ($O_0$), whereas on the test cases which contain trend ($\{T, S1, S2\}$) the best option seems to be $O_2$, which only includes only the increasing counts until the peak of the outbreak is reached ($O_2$).
\subsection{Comparison to the Statistical Surveillance Baselines}
\label{sec:eval_compare}

Considering the results of the previous experiment, we have chosen to evaluate the $p$-value and the standard fusion approach with a window size of $1$, the adaption of the labeling $O_3$ and including the mean. In order to draw conclusions, we have evaluated the underlying statistical methods itself which serve as a baseline.

The results for the average rank are represented in Table~\ref{tab:best_dr}. Here, we can observe that $p$-value fusion achieves the best rating across all test cases. In contrast, the performance of the standard fusion approach is often worse than the underlying statistical algorithms. 
In line with \citet{fusion1} and \citet{fusion2}, we can observe an improvement of the standard fusion approach on the time series without trend and seasonality. However, this improvement is not consistent for all compared test cases, resulting only in an average rank of $3.0$ while our proposed $p$-value approach always achieves the best result. 
Indeed, the ability to detect outbreaks with the standard fusion approach is reduced since it is based on the output of the statistical algorithms given a particular pre-defined significance level $\alpha$ for them. This limits the information about sudden changes encapsulated in the training data which makes it pretty difficult for the ML algorithm to identify valuable patterns. 
A closer examination reveals that trend and seasonality has an impact on the evaluated stacking approaches. In particular, by learning over the baseline data of time series which include trend, the learner is fed with observations which are not representative for the future (evaluation data) due to the changed circumstances. Moreover, the learning algorithm usually assumes that the instances are considered to be independent and identically distributed in the learning data set, not allowing to capture concept drift. Our proposed approaches are not designed to adjust to these settings but we believe that further investigations on the influence of trend and seasonality and how they can be handled is an interesting avenue for future work.

\added{Furthermore, we have evaluated the approaches with respect to the number of cases per outbreak. In contrast to the previous experiments, where the value for the parameter $k$ (used to define the number of cases per outbreak) was randomly drawn between $1$ and $10$, we have fixed this parameter to a particular value for all time series of the $42$ test cases. The results for the measure $dAUC_{1\%}$ across the $42$ test cases with a fixed value for the parameter $k$ is visualized as box plots, representing minimum, first quantile, mean, third quantile and maximum, in Figure~\ref{figure:DRandTPR}.}
In addition to $dAUC_{1\%}$, we include the analysis of the $pAUC_{1\%}$ measure and compare to the original labeling $O_0$ in order to further investigate the effect of the labeling on detection rate and true positive rate.

As the cases per outbreak increases all methods are more likely to obtain a better performance.
\added{While the C1, C2, C3 and RKI method achieve comparable results across all outbreak sizes, we are surprised to observe that the Bayes method has a better performance in case of larger outbreaks. This contradicts our expectation that the RKI method should obtain the best results across these methods since the Poisson assumption was specifically used  to generate the synthetic data.
} 
Regarding the $p$-value fusion approaches, the results confirm the better overall performance across all outbreak sizes while the performance of the standard fusion approach gets worse compared to the other methods with an increasing number of cases per outbreak. This gives further evidence that the standard fusion is not ideal.
A closer examination of the graphs for the measures $dAUC_{1\%}$ and $pAUC_{1\%}$ reveals the difference between the adaption of the labeling for the learning. In particular, without adaption the ML algorithm achieves a tremendously better performance for the trade-off between the true positive rate and the false alarm rate. However, this also has an effect on the ability to detect outbreaks as discussed in Section~\ref{sec:adaption_labeling}, yielding a slightly worse result for the measure $dAUC_{1\%}$ than with adapting the labeling.

\section{Conclusions}
\label{section:conclusion}

In this work, we introduced an approach for the fusion of outbreak detection methods using machine learning, more specifically stacking. 
The original idea is to use the \emph{alarm} or \emph{no alarm} prediction of the underlying statistical algorithms as inputs to the learner.
We improved that setup by incorporating the $p$-values instead, which contain more information about the certainty of an event than the simple binary outputs.
In addition, we proposed to add additional information to the learning data and to adapt the labeling of an outbreak in order to improve the ability to detect outbreaks. 
For evaluation, we proposed a measure based on ROC curves which better adapts to the specific need for a very low false alarm rate 
but still considers the trade-off with the detection rate.

Our experimental results on synthetic data show that the fusion of $p$-values improves the performance compared to the underlying statistical algorithms. Contrary to previous work, we could also observe that simple fusion of binary outputs 
using stacking does not always lead to an improvement. By incorporating additional information to the learning data, 
more specifically the mean count of the previous observations and the previous outputs of the statistical methods,
the machine learning algorithm is able to capture more reliable patterns to detect outbreaks. Furthermore, the labeling of an outbreak has an influence on the performance for the classification algorithm to detect outbreaks. By setting the focus on the peak of an outbreak during the learning process, a better performance to detect sudden changes can be achieved.

\added{
The effectiveness of the proposed method has still to be confirmed on real data. Nevertheless, our results suggest that $p$-value stacking is generally well-suited for combining the outcomes of established methods for outbreak detection with only a low risk of decreasing performance.\com{the risk argument is maybe too ambitious. perhaps "for robustly combining"}
Moreover, stacking allows to enrich the detection by additional signals and sources of information in a highly flexible way.  However, a major challenge remains the treatment of the outbreak annotations during training,  since these labels are  inherently non-binary (endemic vs. epidemic) and additionally noisy and unreliable for real data. 
}




\begin{acks}
This work was supported by the Innovation Committee of the Federal Joint Committee (G-BA) [ESEG project, grant number 01VSF17034]
\end{acks}

\bibliographystyle{ACM-Reference-Format}



\end{document}